\documentclass[11pt]{article}

\usepackage[final]{acl}

\usepackage{times}
\usepackage{latexsym}

\usepackage[T1]{fontenc}
\usepackage[utf8]{inputenc}

\usepackage{microtype}

\usepackage{inconsolata}

\usepackage{graphicx}

\usepackage{booktabs}
\usepackage{amsmath}
\usepackage{CJKutf8}
\usepackage{multirow}
\usepackage{array}
\usepackage{enumitem}

\title{Language Exposure Matters: Distribution and Order in Multilingual BabyLM Pre-training}
\author{JiWoo Hwang$^{1,2}$ \qquad Lukas Edman$^{1,2}$ \qquad Alexander Fraser$^{1,2,3}$ \vspace{.2cm}\\
 $^{1}$School of Computation, Information and Technology, TU Munich \\
$^{2}$Munich Center for Machine Learning \\
 $^{3}$Munich Data Science Institute \\
\vspace{.1cm} {\tt \small jiwoo.hwang@tum.de, lukas.edman@tum.de}
}

\begin{document}
\maketitle
\begin{abstract}
We investigate two aspects of multilingual pre-training with limited data, namely how language exposure is distributed across English, Dutch, and Chinese, and how that exposure is ordered during training. In particular, we compare balanced multilingual exposure to language-dominant exposure, where one language is exposed with more training data than the others. We also compare \textit{mixed training}, where multilingual training data is used throughout training, with \textit{curriculum training}, where monolingual exposure of the dominant language is followed by balanced multilingual exposure. Our results show that multilingual BabyLM performance varies with both the amount and order of language exposure. Mixed training outperforms curriculum learning across all 15 matched configurations, indicating that simultaneous multilingual exposure is more effective than staged exposure in our setting. Several language-dominant mixed training configurations achieve higher scores than the balanced configuration, suggesting that a balanced distribution is not necessarily the highest-performing configuration in our setting. Interestingly, the Dutch-primary configurations achieve the best overall performance.

\end{abstract}

\section{Introduction}
While the performance of multilingual pre-trained models has improved over the recent years, billions of tokens are required to achieve these results. In the context of the BabyLM Challenge \cite{choshen2026babylmturns4goes}, the focus lies on the learning abilities of language models from a limited amount of training data, thereby shifting attention from scaling pre-training toward efficient language learning. For the multilingual track, this brings attention to how data should be distributed among languages, as well as the order of language exposure. 

\begin{figure}
    \centering
    \includegraphics[width=\linewidth]{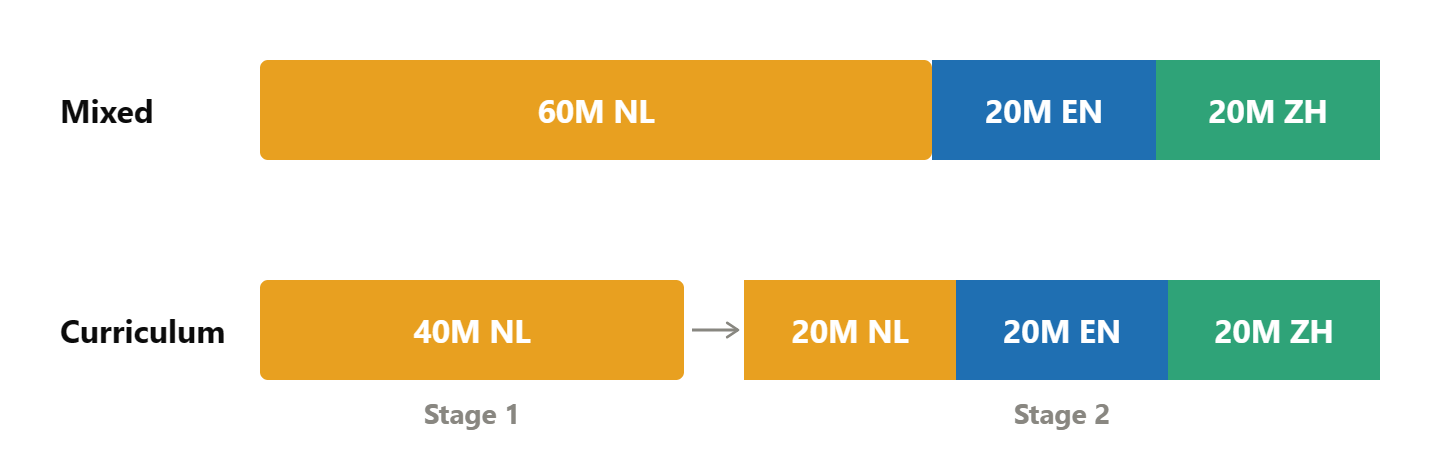}
    \caption{Mixed and curriculum training for the Dutch-primary 60/20/20 configuration. Mixed training shuffles all 100M tokens in a single stage, whereas curriculum training first uses 40M Dutch tokens and then a balanced multilingual stage containing 20M tokens from each language. For this configuration, mixed training achieves a higher multilingual average than curriculum training and produces the best overall result in our experiments.}
    \label{fig:example_mixed_curr}
\end{figure}

In this paper, we address these questions in a controlled setting of multilingual exposure with English, Dutch and Chinese. We compare balanced exposure among the three languages with a setup where one language is exposed more than others. We also compare two training schemes: \textit{mixed} and \textit{curriculum} training. In mixed training, the full multilingual dataset is used throughout training. In curriculum training, the same dataset is used, however split into two stages: the model first trains on a monolingual primary language subset, then continues on a balanced multilingual subset containing all three languages. Our contributions are:

\begin{itemize}[noitemsep, topsep=0pt]
    \item We examine the effect of the amount of language exposure in multilingual BabyLM pre-training by comparing balanced exposure with language-dominant exposure across English, Dutch, and Chinese.
    \item We explore the effect of exposure order by comparing mixed training with a two-stage curriculum learning strategy, in which monolingual primary language exposure is followed by balanced multilingual exposure.
    \item We investigate whether strong overall multilingual performance requires balanced training by comparing settings where each language is dominant.
\end{itemize}
A natural baseline for multilingual training is a balanced distribution across languages, however, our results show that this assumption does not hold in our setting\footnote{Code and experimental resources are available at \url{https://github.com/jinx2321/babylm26}.}. 

\section{Related Work}
\paragraph{Multilingual data distribution.} One of the issues in multilingual learning is whether training should be balanced across different languages. Prior research on multilingual machine translation has demonstrated that the language sampling distribution has a considerable impact on overall performance. \citet{wang2020balancing} show that learning how to rebalance training data across languages can improve performance across languages.

Although balanced exposure to multiple languages serves as a natural baseline, previous work has shown that imbalance is occasionally beneficial for cross-lingual transfer. \citet{schafer2024role} investigate controlled experiments using cloned languages that separate the effects of imbalance from linguistic factors. They show that the dominance of one language positively affects the performance of other minority languages and results in better cross-lingual alignment, an effect that is more prominent in larger models and with more training time. We address a similar problem using the BabyLM framework but with real languages.

\paragraph{Exposure order.} \citet{bengio2009curriculum} introduced curriculum learning, which sorts training examples in a meaningful order, from simple to complex or from one domain to another, in order to boost learning efficiency. In the case of multilingual settings, the ordering may define the moment the model forms shared multilingual representations early or first specializes in one language. Previous research on few-shot cross-lingual transfer has demonstrated that the mixed multilingual training procedure outperforms staged target adaptation \cite{xu2022que}.

Our comparison between mixed and curriculum training is also motivated by bilingual language acquisition, in which both the amount of exposure and the time at which that exposure takes place are important. A distinction is made between simultaneous bilingual language acquisition, in which more than one language is learned from an early age, and sequential bilingual language acquisition, in which another language is acquired later \cite{johnson1989critical}. Although our experimental setup cannot be considered as direct representations of bilinguals, this distinction provides a useful motivation for comparing simultaneous multilingual exposure with staged curriculum learning. 

\section{Experimental Setup}
The detailed training setup can be found in Appendix~\ref{sec:appendix_a}.
\subsection{Data}
\label{sec:data}
The training corpus is constructed from Wikipedia dumps\footnote{\url{https://huggingface.co/datasets/wikimedia/wikipedia}} for the three languages: English, Dutch, and Chinese in the BabyLM 2026 multilingual track. Wikipedia dumps are used in order to keep the source domain as comparable as possible across the languages, since the BabyLM corpora contain different mixtures of domains. As per the constraint of the track, each configuration contains 100M tokens, where the number of tokens is calculated based on the byte premium \cite{arnett2024bit} of each language.

We use six data configurations, where each configuration has different proportions allocated to each language. The baseline configuration contains an equal number of tokens in all three languages. The five language-dominant configurations allocate the primary language with a higher percentage while the rest is divided equally among the other two languages. We have chosen primary language percentages of 50\%, 60\%, 70\%, 80\%, and 90\%, resulting in five different configurations as shown in Table~\ref{tab:configs}.

\begin{table}[h]
\centering
\small
\begin{tabular}{lrrr}
\toprule
\textbf{Configuration} & \textbf{Primary} & \textbf{Other 1} & \textbf{Other 2} \\
\midrule
Balanced       & 33\% & 33\% & 33\% \\
50/25/25       & 50\% & 25\% & 25\% \\
60/20/20       & 60\% & 20\% & 20\% \\
70/15/15       & 70\% & 15\% & 15\% \\
80/10/10       & 80\% & 10\% & 10\% \\
90/5/5         & 90\% &  5\% &  5\% \\
\bottomrule
\end{tabular}
\caption{Data proportion configurations. Each language-dominant
configuration is instantiated with English, Dutch, and Chinese as the
primary language respectively.}
\label{tab:configs}
\end{table}

For each language-dominant split and primary-language pairing, we generate two further subsets for curriculum training (see Section~\ref{sec:training}): a Stage 1 subset containing only primary-language text and a Stage 2 subset containing the remaining primary-language text together with all non-primary-language text, shuffled randomly. The split is constructed so that each language contributes an equal number of tokens in Stage 2. For instance, in the 60/20/20 configuration, Stage 1 contains 40M tokens from the primary language, while Stage 2 contains the remaining 60M tokens, with 20M tokens from each language, as shown in Figure~\ref{fig:example_mixed_curr}.

\subsection{Model}
We train a decoder-only transformer from scratch using the Qwen3 architecture \cite{yang2025qwen3}, which results in ca. 110M parameters (768 hidden dimension, 12 layers, 12 attention heads, 2048 feed-forward dimension, 1024 context length). We train a separate byte-level BPE tokenizer for each data configuration, using the training data of the configuration as the corpus for tokenization. This guarantees that the language distribution utilized for tokenization is equal to the distribution seen during the pre-training process. 

\subsection{Training Strategies and Evaluation}
\label{sec:training}
We compare the mixed and curriculum learning strategies across all five language-dominant configurations, as shown in Figure \ref{fig:example_mixed_curr}. In mixed training, all 100M tokens from the corpus are shuffled in random order and used during the 10-epoch training. The model is exposed to all three languages throughout all training iterations. Curriculum training involves two consecutive training stages using the Stage 1 and Stage 2 subsets from Section~\ref{sec:data}. In Stage 1, the model trains on the subset that contains only the primary language for 10 epochs, and then Stage 2 continues training on the multilingual balanced subset for another 10 epochs, initialized from the Stage 1 checkpoint, including the optimizer and learning rate scheduler states. Stage 1 and Stage 2 together cover the whole 100M-token dataset configuration. Thus, each language is seen the same number of times across both strategies and only the order of exposure differs.

For all models, the official BabyLM evaluation pipeline \cite{choshen2026babylmturns4goes} was used for evaluation, which contains zero-shot and fine-tuning tasks in English, Dutch, and Chinese. The zero-shot tasks include BLiMP \cite{warstadt2020blimp}, GlobalPIQA \cite{chang2025global}, HellaSwag \cite{zellers2019hellaswag}, MultiBLiMP \cite{jumelet2026multiblimp}, Winogrande \cite{sakaguchi2021winogrande}, XStoryCloze \cite{lin2022few}, and XCOMPS \cite{he2025xcomps}, while fine-tuning evaluation cover tasks such as ARC \cite{clark2018think}, Belebele \cite{bandarkar2024belebele}, BMLAMA \cite{qi2023cross}, MNLI \cite{williams2018broad}, SIB-200 \cite{adelani2024sib}, TruthfulQA \cite{lin2022truthfulqa}, XNLI \cite{conneau2018xnli}, and INCLUDE \cite{romanou2025include}. The zero-shot average, fine-tuning average, and overall average scores of each language are reported for all models.

\section{Results and Discussion}
The full results can be found in Appendix~\ref{sec:appendix_b}.
\subsection{Impact of Language Exposure Proportion}
\begin{figure}[h!]
\centering
\includegraphics[width=\linewidth]{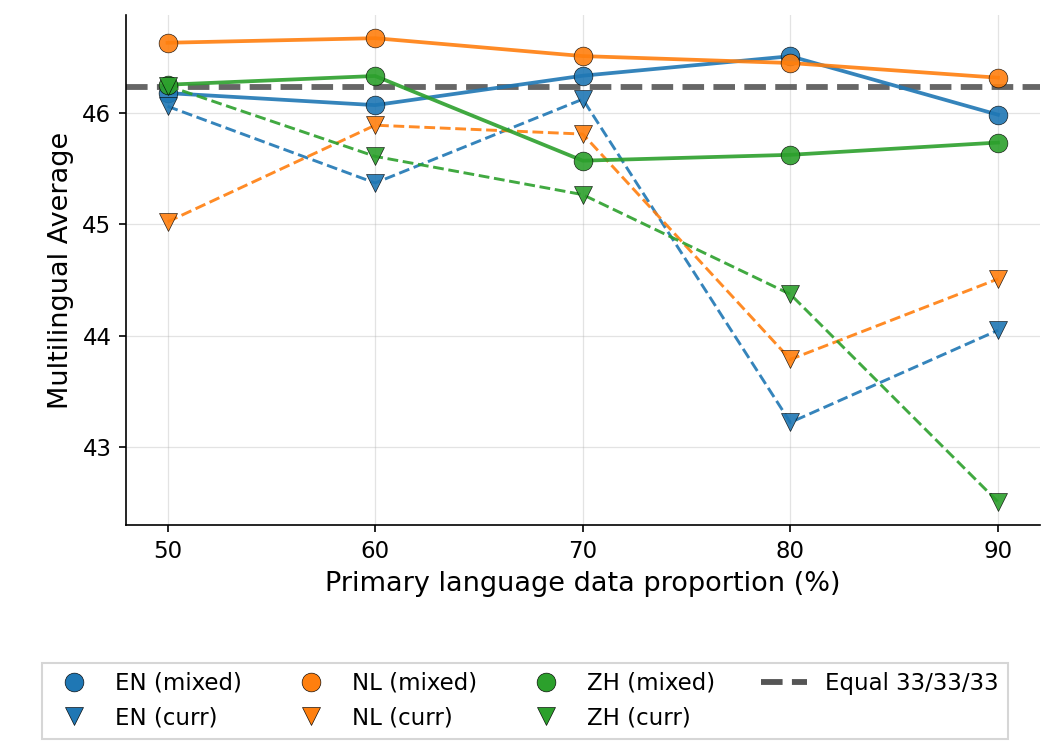}
\caption{Multilingual average performance across primary-language data proportions, grouped by the language designated as primary and by training strategy. Each point averages performance across English, Dutch, and Chinese, while the dashed horizontal line marks the balanced 33/33/33 baseline.}
\label{fig:multi-avg}
\end{figure}
\begin{figure*}[h!]
\centering
\includegraphics[width=\linewidth]{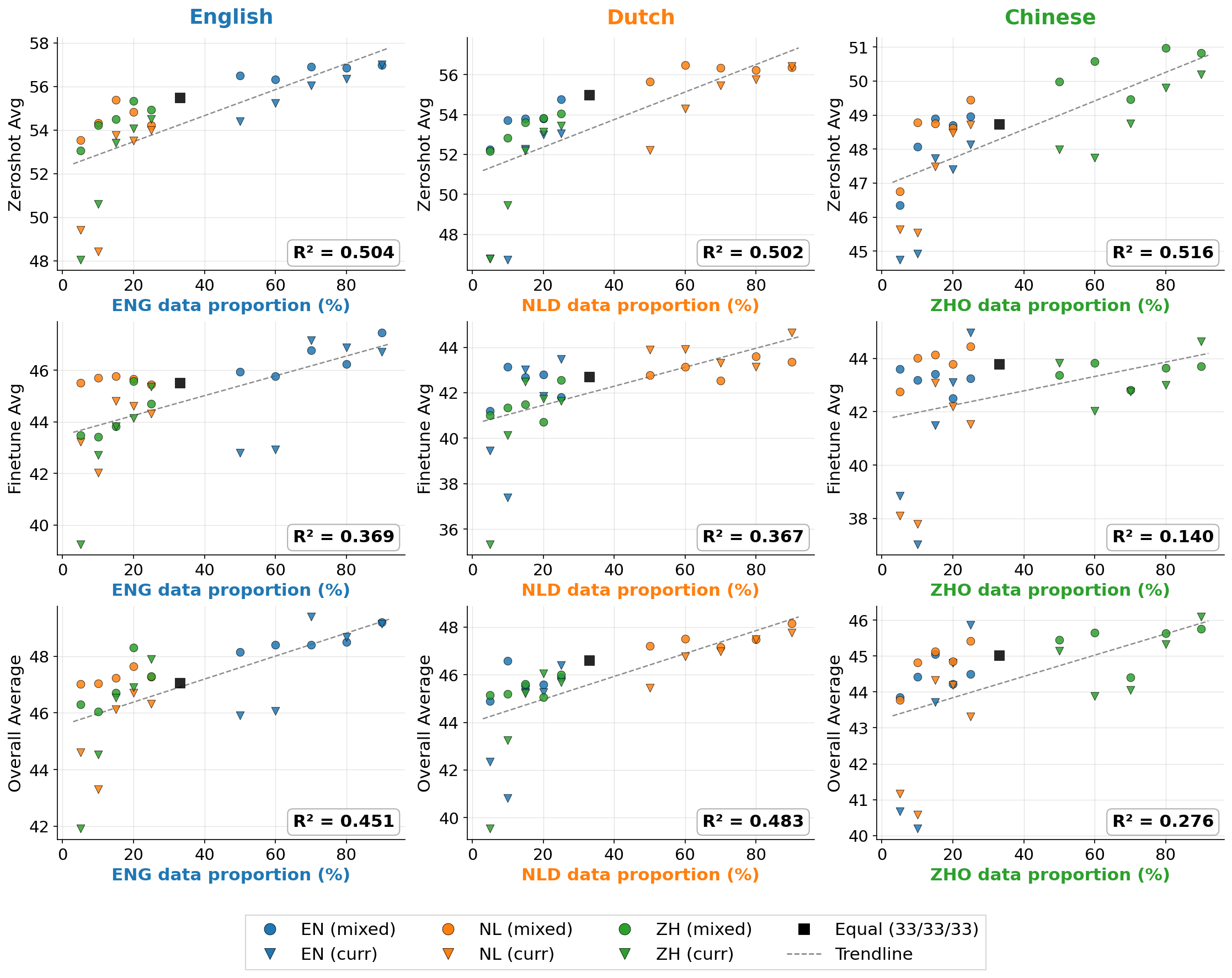}
\caption{Relationship between language-specific data proportion and evaluation performance. Columns show English, Dutch, and Chinese data proportions, while rows show zero-shot, finetuned, and overall average performance for each language. Dashed lines indicate linear trends across all models, with $R^2$ reported in each panel. The black square represents the balanced 33/33/33 baseline.}
\label{fig:9-panel}
\end{figure*}

We analyze the effect of language exposure from two complementary perspectives. Figure \ref{fig:multi-avg} compares complete data-allocation configurations using their multilingual average, showing whether language-dominant exposure improves overall performance relative to the balanced baseline. Figure \ref{fig:9-panel} instead examines each language separately by relating its proportion in the training data to performance in that same language.

We first compare the language-dominant configurations with the balanced 33/33/33 mixed training baseline, which obtains a multilingual average of 46.23. For English-primary mixed training, the 70/15/15 and 80/10/10 configurations outperform the balanced baseline, with performance peaking at 46.51 for 80/10/10. 

Dutch-primary mixed training performs more consistently: all five Dutch-primary configurations outperform the balanced baseline. The highest multilingual average is 46.67 at 60/20/20, an improvement of 0.44 points over the baseline and the best overall result among all configurations. Chinese-primary training provides smaller benefits. The 50/25/25 and 60/20/20 configurations slightly exceed the baseline, reaching a maximum of 46.33 at 60/20/20, whereas configurations with higher Chinese proportions fall below it.

These comparisons show that the balanced configuration is not the highest-scoring configuration in this setting, although the benefit of language dominance depends on both the primary language and the degree of imbalance. English-primary performance peaks at 80/10/10, while Dutch-primary and Chinese-primary performance peak at 60/20/20. At 90/5/5, only the Dutch-primary model remains above the balanced baseline, indicating that extreme imbalance is generally less effective.

This finding is important since a balanced distribution is a natural baseline when all three evaluation languages are weighted equally. Nevertheless, for each primary language, at least one language-dominant mixed training configuration outperforms the balanced model. The advantage is therefore not limited to English dominance as the Dutch-primary 60/20/20 configuration achieves the best overall result, while the Chinese-primary 50/25/25 and 60/20/20 configurations also surpass the balanced baseline. At the same time, the weaker results under the most extreme allocations suggest a trade-off between increasing exposure to the primary language and maintaining sufficient exposure to the other languages.

\subsection{Mixed vs. Curriculum Learning}
Next, we compare mixed training with the curriculum learning strategy. As shown in Figure~\ref{fig:multi-avg}, mixed training achieves a higher multilingual average than curriculum training in all 15 matched language-dominant configurations. The difference is generally larger at high levels of dominance, although it does not increase monotonically. For example, at 80/10/10, mixed training outperforms curriculum training by 3.29 points in the English-primary setting and by 2.66 points in the Dutch-primary setting.

The major weakness of curriculum training is its decline in performance of non-primary languages under high dominance conditions. In the English-primary 80/10/10 case, the Dutch and Chinese average scores of mixed training models are 46.59 and 44.42, respectively, whereas the corresponding figures for curriculum models are 40.81 and 40.19. This suggests that the large monolingual first stage encourages specialization on the primary language, and the subsequent multilingual stage does not fully compensate for this, consistent with findings that simultaneous multilingual exposure outperforms staged adaptation~\cite{xu2022que}.

\subsection{Language-Specific Effects of Exposure}

Concerning the language-specific results in Figure~\ref{fig:9-panel}, the clearest relationship between data proportion and performance appears in zero-shot evaluation. The linear trends are positive for all three languages, with $R^2$ values of 0.504 for English, 0.502 for Dutch, and 0.516 for Chinese. These similar values indicate that the proportion of a language in the pre-training data has a moderate relationship with zero-shot performance in that same language.

The relationship is weaker for finetuned performance. English and Dutch show similar $R^2$ values of 0.369 and 0.367, respectively, whereas the relationship is much weaker for Chinese, with $R^2=0.140$. This difference is also visible in the trend lines as English and Dutch finetuned performance generally increases with additional same-language exposure, while Chinese finetuned scores vary less consistently with the proportion of Chinese data.

The language-specific overall averages combine the zero-shot and finetuned results. The relationship is strongest for Dutch ($R^2=0.483$), followed by English ($R^2=0.451$), and is weaker for Chinese ($R^2=0.276$). Overall, Figure~\ref{fig:9-panel} shows that same-language data proportion is most consistently related to zero-shot performance. The dispersion around the trend lines also shows that data proportion alone does not fully account for the differences among configurations.

\section{Conclusion}
Our work studies the impact of language distributions and order of exposure on BabyLM multilingual pre-training with English, Dutch, and Chinese on a 100M-token budget. By comparing balanced and language-dominant data distributions with mixed and curriculum training strategies, we examined how both the amount and order of multilingual exposure shape evaluation performance.

We find that mixed training outperforms curriculum learning. Although the size of the difference does not increase monotonically, the gap is generally larger under highly imbalanced allocations, where curriculum training shows substantial declines in non-primary-language performance. Hence, simultaneous multilingual training should be preferred to sequential monolingual specialization and multilingual adaptation.

In addition, perhaps more surprisingly, pre-training on balanced data is not the optimal approach in our experiments. For mixed training, even though we consider a balanced configuration as the baseline, a certain level of imbalance in favor of one language improves the performance of all three primary languages, suggesting that the effect of imbalance is not language-specific. However, the advantage decreases for extremely large imbalances, indicating a trade-off between strengthening the dominant language and preserving cross-lingual coverage. However, since we evaluate multiple language-dominant configurations against a single balanced configuration and conduct only one run per setting, these results should be interpreted descriptively rather than as evidence that imbalanced training is generally preferable to balanced training.

\section*{Limitations}
Our study is limited to only three languages (English, Dutch, and Chinese) within one domain (Wikipedia), thus it is uncertain whether our findings will apply to other languages or domains. All experiments use a fixed size of 100M tokens and a single model architecture (Qwen3). Results may vary for other scales and other architectures. The training procedure uses a separate tokenizer for every experiment based on the particular training dataset. Hence, some variations may be due to tokenizer settings in addition to differences in the data distribution. However, this choice was made in order to ensure that the language distribution for tokenization and model pre-training is the same. Moreover, due to limited computational resources, we conduct only one training procedure per configuration with a fixed random seed. Therefore, we do not estimate variance across training runs. The curriculum learning strategy we explore is applied using a fixed protocol with two stages, thus different protocols such as gradual mixing or additional stages may produce different results.

\section*{Acknowledgments}
The authors gratefully acknowledge the scientific support and HPC resources provided by the Erlangen National High Performance Computing Center (NHR@FAU) of the Friedrich-Alexander-Universität Erlangen-Nürnberg (FAU) under the NHR project b279bb. NHR funding is provided by federal and Bavarian state authorities. NHR@FAU hardware is partially funded by the German Research Foundation (DFG) – 440719683.

This work was co-funded by the European Union (ERC, EPICAL, 101141712). Views and opinions expressed are however those of the author(s) only and do not necessarily reflect those of the European Union or the European Research Council. Neither the European Union nor the granting authority can be held responsible for them.

\bibliography{custom}

\appendix

\section{Training Details}
\label{sec:appendix_a}
Table~\ref{tab:hyperparams} summarizes the shared training setup used for all models. All configurations use the same model size, optimization settings, context length, and random seed.

\begin{table}[h!]
\centering
\small
\begin{tabular}{lr}
\toprule
\textbf{Setting} & \textbf{Value} \\
\midrule
Model size              & $\sim$110M parameters \\
Tokenizer               & Byte-level BPE \\
Vocabulary size         & 32{,}000 \\
Context length          & 1024 \\
\midrule
Optimizer               & AdamW \\
Learning rate           & $6 \times 10^{-4}$ \\
$(\beta_1, \beta_2)$    & $(0.9,\ 0.95)$ \\
Weight decay            & 0.1 \\
LR schedule             & Cosine with warmup \\
Warmup                  & 5\% of steps \\
Batch size              & 128 sequences \\
\midrule
Epochs (mixed)          & 10 over 100M tokens\\
Epochs (curriculum)     & 10 per subset \\
Random seed             & 42 \\
Hardware                & 1$\times$ NVIDIA H100 \\
\bottomrule
\end{tabular}
\caption{Training setup used for all models.}
\label{tab:hyperparams}
\end{table}

\section{Full Results}
\label{sec:appendix_b}
Table~\ref{tab:full_results} reports the complete results for all 31 models. We include zero-shot, finetuned, language-specific average, and multilingual average scores for each configuration to show how language distribution and training strategy affect both individual languages and overall multilingual performance.
\begin{table*}[h!]
\centering
\footnotesize
\setlength{\tabcolsep}{5pt}
\begin{tabular}{llrrr rrr rrr r}
\toprule
\multicolumn{2}{l}{\textbf{Model}} & \multicolumn{3}{c}{\textbf{English}} & \multicolumn{3}{c}{\textbf{Dutch}} & \multicolumn{3}{c}{\textbf{Chinese}} & \textbf{Multi} \\
\cmidrule(lr){3-5}\cmidrule(lr){6-8}\cmidrule(lr){9-11}
\textbf{Config} & \textbf{Strategy} & ZS & FT & Avg & ZS & FT & Avg & ZS & FT & Avg & Avg \\
\midrule
\multicolumn{12}{l}{Balanced baseline} \\
  33/33/33 & Mixed & 55.48 & 45.52 & 47.06 & 54.99 & 42.70 & 46.61 & 48.73 & 43.78 & 45.03 & 46.23 \\
\midrule
\multicolumn{12}{l}{English-primary} \\
  50/25/25 & Mixed & 56.50 & 45.94 & 48.15 & 54.77 & 41.80 & 45.89 & 48.96 & 43.25 & 44.50 & 46.18 \\
   & Curr & 54.39 & 42.79 & 45.91 & 53.05 & 43.47 & 46.39 & 48.13 & \textbf{44.97} & 45.87 & 46.06 \\
  60/20/20 & Mixed & 56.31 & 45.78 & 48.40 & 53.80 & 42.80 & 45.59 & 48.70 & 42.50 & 44.22 & 46.07 \\
   & Curr & 55.23 & 42.93 & 46.06 & 52.99 & 41.86 & 45.25 & 47.41 & 43.11 & 44.80 & 45.37 \\
  70/15/15 & Mixed & 56.91 & 46.78 & 48.41 & 53.79 & 42.68 & 45.53 & 48.90 & 43.41 & 45.06 & 46.33 \\
   & Curr & 56.04 & 47.17 & \textbf{49.39} & 52.26 & 43.03 & 45.26 & 47.73 & 41.49 & 43.72 & 46.12 \\
  80/10/10 & Mixed & 56.87 & 46.25 & 48.51 & 53.72 & 43.13 & 46.59 & 48.07 & 43.19 & 44.42 & 46.51 \\
   & Curr & 56.35 & 46.88 & 48.67 & 46.72 & 37.40 & 40.81 & 44.92 & 37.04 & 40.19 & 43.22 \\
  90/5/5 & Mixed & 56.98 & \textbf{47.47} & 49.21 & 52.24 & 41.20 & 44.89 & 46.34 & 43.60 & 43.85 & 45.98 \\
   & Curr & \textbf{57.00} & 46.71 & 49.14 & 46.77 & 39.44 & 42.34 & 44.75 & 38.84 & 40.68 & 44.05 \\
\midrule
\multicolumn{12}{l}{Dutch-primary} \\
  50/25/25 & Mixed & 54.23 & 45.46 & 47.27 & 55.65 & 42.76 & 47.20 & 49.45 & 44.46 & 45.42 & 46.63 \\
   & Curr & 54.00 & 44.31 & 46.32 & 52.23 & 43.88 & 45.44 & 48.72 & 41.54 & 43.31 & 45.02 \\
  60/20/20 & Mixed & 54.82 & 45.66 & 47.65 & \textbf{56.48} & 43.14 & 47.51 & 48.62 & 43.79 & 44.85 & \textbf{46.67} \\
   & Curr & 53.51 & 44.61 & 46.71 & 54.29 & 43.91 & 46.77 & 48.48 & 42.20 & 44.19 & 45.89 \\
  70/15/15 & Mixed & 55.40 & 45.77 & 47.24 & 56.35 & 42.52 & 47.16 & 48.75 & 44.14 & 45.13 & 46.51 \\
   & Curr & 53.77 & 44.81 & 46.13 & 55.45 & 43.30 & 46.98 & 47.48 & 43.09 & 44.33 & 45.81 \\
  80/10/10 & Mixed & 54.32 & 45.70 & 47.05 & 56.24 & 43.61 & 47.48 & 48.78 & 44.02 & 44.82 & 46.45 \\
   & Curr & 48.42 & 42.02 & 43.31 & 55.75 & 43.13 & 47.48 & 45.55 & 37.79 & 40.58 & 43.79 \\
  90/5/5 & Mixed & 53.54 & 45.51 & 47.01 & 56.37 & 43.37 & \textbf{48.16} & 46.77 & 42.75 & 43.77 & 46.32 \\
   & Curr & 49.42 & 43.22 & 44.61 & 56.42 & \textbf{44.66} & 47.76 & 45.63 & 38.10 & 41.17 & 44.51 \\
\midrule
\multicolumn{12}{l}{Chinese-primary} \\
  50/25/25 & Mixed & 54.93 & 44.70 & 47.30 & 54.05 & 42.56 & 46.00 & 49.98 & 43.37 & 45.46 & 46.25 \\
   & Curr & 54.49 & 45.38 & 47.90 & 53.43 & 41.65 & 45.69 & 47.98 & 43.83 & 45.14 & 46.24 \\
  60/20/20 & Mixed & 55.33 & 45.57 & 48.30 & 53.83 & 40.70 & 45.05 & 50.58 & 43.82 & 45.65 & 46.33 \\
   & Curr & 54.06 & 44.14 & 46.90 & 53.14 & 41.73 & 46.05 & 47.74 & 42.02 & 43.88 & 45.61 \\
  70/15/15 & Mixed & 54.49 & 43.83 & 46.70 & 53.61 & 41.49 & 45.61 & 49.46 & 42.81 & 44.40 & 45.57 \\
   & Curr & 53.40 & 43.83 & 46.54 & 52.20 & 42.49 & 45.22 & 48.75 & 42.74 & 44.05 & 45.27 \\
  80/10/10 & Mixed & 54.21 & 43.41 & 46.05 & 52.83 & 41.35 & 45.19 & \textbf{50.97} & 43.65 & 45.63 & 45.62 \\
   & Curr & 50.61 & 42.71 & 44.54 & 49.47 & 40.13 & 43.25 & 49.80 & 42.99 & 45.33 & 44.37 \\
  90/5/5 & Mixed & 53.06 & 43.48 & 46.30 & 52.17 & 41.02 & 45.14 & 50.82 & 43.70 & 45.75 & 45.73 \\
   & Curr & 48.05 & 39.25 & 41.91 & 46.77 & 35.33 & 39.53 & 50.19 & 44.64 & \textbf{46.09} & 42.51 \\
\bottomrule
\end{tabular}
\caption{Full results for all 31 models. ZS = zero-shot average, FT = finetune average, Avg = combined average for that language, Multi = multilingual average across all three languages. Bold indicates the best score in each column.}
\label{tab:full_results}
\end{table*}

\end{document}